# Printed Arabic Text Recognition using Linear and Nonlinear Regression

Ashraf A. Shahin[1,2]

[1]College of Computer and Information Sciences,
Al Imam Mohammad Ibn Saud Islamic University (IMSIU)
Riyadh, Kingdom of Saudi Arabia
[2]Department of Computer and Information Sciences, Institute of Statistical Studies & Research,
Cairo University,
Cairo, Egypt

*Abstract*—**Arabic language is one of the most popular languages in the world. Hundreds of millions of people in many countries around the world speak Arabic as their native speaking. However, due to complexity of Arabic language, recognition of printed and handwritten Arabic text remained untouched for a very long time compared with English and Chinese. Although, in the last few years, significant number of researches has been done in recognizing printed and handwritten Arabic text, it stills an open research field due to cursive nature of Arabic script. This paper proposes automatic printed Arabic text recognition technique based on linear and ellipse regression techniques. After collecting all possible forms of each character, unique code is generated to represent each character form. Each code contains a sequence of lines and ellipses. To recognize fonts, a unique list of codes is identified to be used as a fingerprint of font. The proposed technique has been evaluated using over 14000 different Arabic words with different fonts and experimental results show that average recognition rate of the proposed technique is 86%.**

## I. INTRODUCTION

Optical Character Recognition (OCR) is a process of analyzing images of printed or handwritten text to translate character image into a machine editable format [1]. Printed character recognition has been extensively used in many areas especially after popularity of electronic media, which increases necessity of converting printed text into the new electronic media.

Several recognition techniques have been proposed recently to recognize printed Arabic characters [2, 3, 4, 5, 7]. Nevertheless, the problem of recognizing printed Arabic characters still is an active area of research and still has many challenges. Arabic text has many characteristics that make the process of recognizing printed Arabic text as a difficult task. These characteristics include [2, 3]:

- Arabic text is cursive script and its characters are connected even in machine printed documents (see Fig. 1).

- Neighboring characters in Arabic script usually overlapped (see Fig. 2), which increases the difficulty of isolating characters.

- Several characters in Arabic script have the same form and the difference is one or more dots in different locations (see Fig. 3).

- Character form may vary between fonts (see Fig. 4). Therefore, the large number of existing Arabic fonts increases the difficulty of recognition task.

- Arabic text has 28 characters and 10 numerals. As shown in Table 1, each character has up to four forms depend on its position in the word (isolated, beginning, middle, and end). Therefore, it is expected that there are 120 different character forms in each font after adding the new character ( ﻻ ), which is created by writing ALIFON ( ﺍ ) after LAMON ( ﻝ ). Although, this number of forms has been mentioned, described and used in many researches [r8][r20][r22][r23], the problem is more larger than this. Unfortunately, each character may have different forms in the same location in the same font. As shown in Fig. 5, BAAON ( ﺏ ) has up to five forms at the beginning of the word in the same font (*Traditional Arabic font*). Its form does not only depend on its neighbor but also depend on the neighbor of its neighbor as in BEMA ( ﺑﻤﺎ ) and BEM ( ﺑﻢ ).

This paper proposes printed Arabic text recognition technique using linear and ellipse regression techniques. Characters are recognized by using Codebook, which contains code for each character form as well as fingerprints to recognize fonts. Characters code and fonts fingerprint are represented as sequences of points, lines, and ellipses. Linear regression is employed to avoid difficulties of representing line segments using ellipses such as infinite major axis and very small minor axis.

Main contribution of the proposed technique is generating codebook contains fingerprint of each font and code for each possible character form without using corpuses. This feature allows the proposed technique to follow rapid generation of new fonts. Moreover, the proposed technique uses regression techniques with ability to fit more than 100,000 pixels in a second [8, 9], which accelerates its recognition speed.

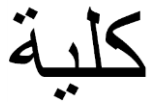

Fig. 1.   Example of connectivity in Arabic script

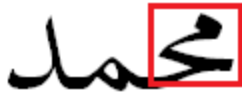

Fig. 2.   Example of neighboring characters overlapping in Arabic script

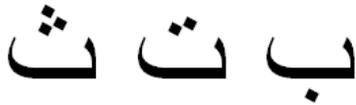

Fig. 3.   Example of different characters with the same form and different number of dots

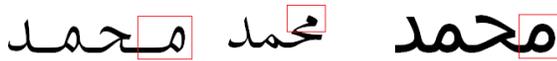

Fig. 4.   Example of different forms of the Same Character MEMON (م) in different fonts

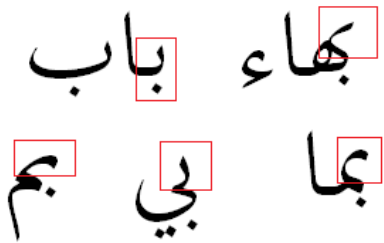

Fig. 5.   Example of different forms of the same character BAAON (ب) in the same location (beginning) in the same font (Traditional Arabic font)

The rest of this paper is organized as follows. Section 2 overviews current related work. Section 3 discusses the proposed technique for recognizing printed Arabic characters. Section 4 explains evaluation methodology and experimental results. Finally, the paper is concluded in Section 5.

## II.   RELATED WORK

Although, there are a large number of printed Arabic characters recognition approaches have been proposed in the last few years, there still needs to enhance recognition rate in Arabic OCR systems. This section overviews some of these approaches.

Rashad et al. [2] have compared between K- Nearest Neighbor (KNN) and Random Forest Tree (RFT) classifiers in recognizing printed Arabic characters. First, global binarization has been used to binarize images. 14 statistical features have been extracted from each character by using horizontal and vertical transitions techniques. Finally, KNN and RFT have been applied to recognize characters. Their experiments show that, although, KNN is faster than RFT in training and testing, RFT performs better than KNN, where recognition rate of RFT is 98% while recognition rate of KNN is 87%.

TABLE I.   DIFFERENT FORMS OF ARABIC CHARACTERS

| isolated | beginning | middle | end |
|---|---|---|---|
| ا | ا | ـا | ـا |
| ب | بـ | ـبـ | ـب |
| ت | تـ | ـتـ | ـت |
| ث | ثـ | ـثـ | ـث |
| ج | جـ | ـجـ | ـج |
| ح | حـ | ـحـ | ـح |
| خ | خـ | ـخـ | ـخ |
| د | د | ـد | ـد |
| ذ | ذ | ـذ | ـذ |
| ر | ر | ـر | ـر |
| ز | ز | ـز | ـز |
| س | سـ | ـسـ | ـس |
| ش | شـ | ـشـ | ـش |
| ص | صـ | ـصـ | ـص |
| ض | ضـ | ـضـ | ـض |
| ط | طـ | ـطـ | ـط |
| ظ | ظـ | ـظـ | ـظ |
| ع | عـ | ـعـ | ـع |
| غ | غـ | ـغـ | ـغ |
| ف | فـ | ـفـ | ـف |
| ق | قـ | ـقـ | ـق |
| ك | كـ | ـكـ | ـك |
| ل | لـ | ـلـ | ـل |
| م | مـ | ـمـ | ـم |
| ن | نـ | ـنـ | ـن |
| ه | هـ | ـهـ | ـه |
| و | و | ـو | ـو |
| ي | يـ | ـيـ | ـي |

Chergui et al. [10] have proposed multiple classifier system (MCS) to recognize Arabic optical characters. The proposed classification engine is based on serial combination of Radial Basic Function (RBF) and set of Adaptive Resonance Theory networks (ART1). RBF-based classifier is used to give a score for the most likely classes based on the first 49 Tchebichef moments, which are extracted after normalizing, aligning, and thinning processes. By using Tchebichef moments, image has been represented with minimum amount of information redundancy. Finally, an adaptive resonance theory network has applied on each group obtained from applying RBF-based classifier. Experimental results have shown that the proposed classification engine outperforms RBF based classifiers and ART1-based classifiers.

Amara et al. [11] have overviewed Arabic OCR using Support Vectors Machines (SVM). Although, SVM has proven its efficiency in different domains among other classification tools, SVM has not been effectively applied in recognizing Arabic characters. The authors have concluded that there are still many challenges face current algorithms that apply SVM in Arabic OCR, such as precision, consistency, and efficiency. The authors have found that best recognition rate has been reached by applying one-against-all technique with Gaussian RFB kernel. Best RFB kernel parameters are determined by using Ten-fold cross validation.

Jiang et al. [12] have proposed small-size printed Arabic text recognition approach based on hidden Markov (HMM) model estimation. Although, applying hidden Markov model has some advantages (such as no pre-segmentation), bad image quality of small-size printed Arabic text makes it difficult to find accurate model boundary. In the proposed approach, state number of HMM has been optimized and bootstrap approach has been modified to improve accuracy of finding model boundary of small-size printed Arabic text with bad image quality. Bootstrap approach has been modified by using some HMMs with different state number and select HMM with the best performance before Viterbi alignment. Their experimental results show that error rate of word recognition is decreased 13.3% and error rate of character recognition is decreased 14%.

Ahmed et al. [13] have employed a special type of recurrent neural network, called bidirectional long short-term memory (BLSTM) networks, to propose segmentation-free optical character recognition system. BLSTM has proven its efficiency in many research areas due to its ability to remember events when there are long time lags between events. However, BLSTM requires pre-segmented training data, and post-processing to transform outputs into label sequences. Therefore, layer called connectionist temporal classification (CTC) has been used with BLSTM to label unsegmented sequences directly. The proposed approach has been evaluated with cursive Urdu and non-cursive English scripts. Although, their experiments show that accuracy of the proposed approach is 99.17% with non-cursive, its accuracy is 88.94% with cursive. Therefore, the proposed approach needs more investigation to enhance it accuracy with cursive scripts.

Accuracy of optical character recognition system influences by graphical entities (e.g., horizontal or vertical edges, symbols, logos) that are exist in printed document image. To overcome this problem, Rani et al. [14] have proposed algorithm to detect such graphical entities. The proposed algorithm detects graphical entities by using Zernike moments and histogram of gradient features and detects horizontal and vertical lines by masking the image with rectangular structuring element. Their experimental outcomes show that accuracy of the proposed algorithm is 97% in detecting graphical entities and 92% in detecting horizontal and vertical lines.

Sarfraz et al. [3] have proposed offline Arabic character recognition system. The proposed system has four stages. In the first stage, text-preprocessing stage, removes isolated pixel and correct drift. Pixel is considered isolated if it does not have any neighboring pixels. Drift is corrected by rotating the image according to the angle with highest number of occurrences between all angles of all lines segments between any pair of black pixels in the image. In the second stage, line and word segmentation are performed by using horizontal and vertical projection. Words are segmented into individual characters by comparing vertical projection profile with fixed threshold. Feature space is built by using moment invariant technique. Finally, characters are recognized by using two different approaches: syntactic approach and a neural network approach. Their experiments show that recognition accuracy of syntactic approach is 89% - 94%, while recognition accuracy of neural network approach is 83%.

Abdi et al. [15] have proposed text-independent Arabic writer identification and verification approach. Beta-elliptic model has been adapted by the proposed approach to construct its own grapheme codebook instead of extracting natural graphemes from a training corpus using segmentation and clustering. The size of the generated codebook is reduced by using feature selection. Feature vectors are extracted using template matching to perform writer identification and verification.

Zagoris et al. [6] have proposed an approach to differentiate between handwritten and machine-printed text. Text image is segmented into blocks. Each block is represented as word vector, which contains local features that are identified using Scale-Invariant Feature Transform. Based on Support Vector Machines, the proposed approach decides whether text block is handwritten, machine printed, or noise by comparing its word vector with codebook.

## III. THE PROPOSED RECOGNITION TECHNIQUE

Architecture of the proposed recognition technique is shown in Fig. 6. In the preprocessing stage, text image is thinned using Zhang-Suen thinning algorithm [16] and segmented into disconnected sub-words (see Fig. 7).

Relations between segments are represented using Freeman code as following. For each segment with a set of pixels $(x_i, y_i), i = 1, 2, .., N$, center point $(x_c, y_c)$ is defined, where

$$x_c = \frac{1}{N} \sum_{i=1}^{N} x_i \; , \quad and \; y_c = \frac{1}{N} \sum_{i=1}^{N} y_i$$

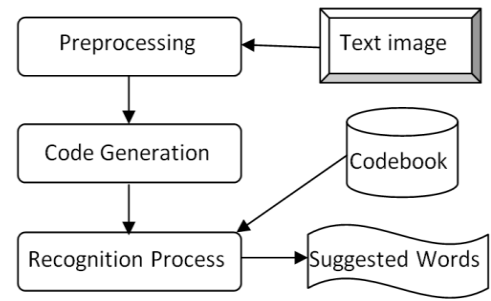

Fig. 6. Architecture of the proposed Recognition Technique

All center points are sorted from left to right and from top to bottom. Directions from each point to the following three points (if exists) are represented by Freeman code, which is shown in Fig. 8.

In the second stage, code is generated using the proposed encoding technique, which is described in subsection *A*. Generated code is compared with characters' code from codebook (described in subsection *C*) using the proposed matching technique (described in subsection *B*) to recognize its characters. Finally, recognized words are introduced.

### A. Proposed Encoding Technique

Arabic script is cursive script. Therefore, Arabic words can be represented by a sequence of lines and curves. The proposed recognition technique generates a sequence of points, lines, and ellipses to represent each sequence of connected characters or sub-characters. Points are used to represent dots that exist in several characters.

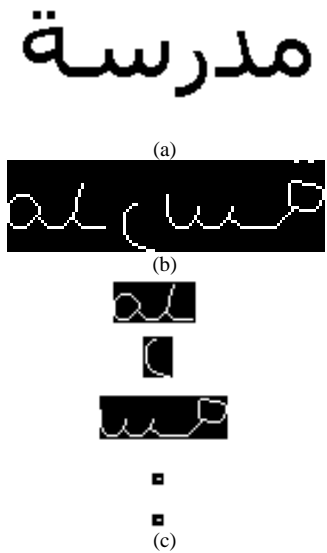

Fig. 7. Thinning and segmenting text image, (a) original word, (b) thinned word, and (c) segmented word into disconnected sub-words

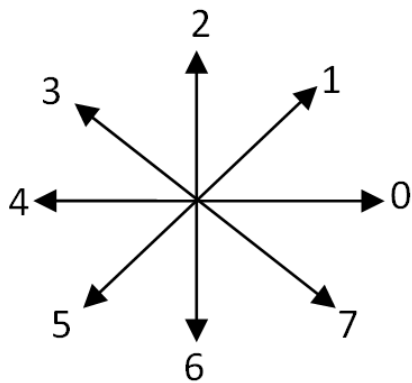

Fig. 8. Freeman code [17]

To collect sequences of connected pixels that can be regressed to lines, it computes the line that passes through the first two pixels. New pixel is added to the list if its distance from the line is less than or equal pre-specified value $\Delta d$

(accuracy factor). After adding new pixel to the list, line is re-calculated to find best line, which fits to all pixels in the new list. If distance between current pixel and the previous line is greater than accuracy factor, algorithm is terminated. If length of line segment that correspond to collected pixels is greater than a pre-specified length, code for this sequence of connected pixels is generated as $(p, \propto, l)$, where $p$ and $\propto$ are parameters of the line ($p$ is distance of the line from origin, and $\propto$ is the angle that the line between closest point on the line and origin makes with the polar axis), and $l$ is the length of line segment.

To find best line that fits to a set of pixels, the proposed recognition technique applies linear regression methodology described in [8]. Best line that fits to a set of pixels $(x_i, y_i), i = 1, 2, .., N$, is identified by the following equation:

$$r \cos(\theta - \alpha) - p = 0, \qquad (1)$$

where

$$\tan 2\alpha = \frac{-2 \sum_{i=1}^{N}(\bar{y} - y_i)(\bar{x} - x_i)}{\sum_{i=1}^{N}[(\bar{y} - y_i)^2 - (\bar{x} - x_i)^2]} \qquad (2)$$

$$p = \bar{x} \cos\alpha + \bar{y} \sin\alpha \qquad (3)$$

$$\bar{x} = \frac{1}{N} \sum_{i=1}^{N} r_i \cos\theta_i$$

$$\bar{y} = \frac{1}{N} \sum_{i=1}^{N} r_i \sin\theta_i$$

$$r_i = \sqrt{x_i^2 + y_i^2}$$

$$\theta_i = \tan^{-1}\left(\frac{y_i}{x_i}\right)$$

Table 2 shows codes of line segments that are detected in sub-words in Fig. 7.

In the previous step, pixels that can be regressed to lines are extracted. In this step, all remaining pixels are clustered to sets such that pixels in each set are connected and can be regressed to ellipse. Each ellipse is described using 6 coefficients as following:

$$F(x, y) = ax^2 + bxy + cy^2 + dx + ey + f = 0 \qquad (4)$$

with

$$b^2 - 4ac < 0 \qquad (5)$$

TABLE II. Examples of Detected Line Segments and Their Codes

| Line segments | code |
|---|---|
| | (27, 54, 12) |
| | (21, 171, 9) |
| | (35, 118, 11) |
| | (13, 124, 6) |
| | (77, 14, 2) |
| | (23, 81, 3) |
| | (51, 118, 11) |
| | (29, 12, 14) |
| | (82, 141, 1) |
| | (98, 89, 10) |
| | (17, 137, 11) |

For each point $(x, y)$, $F(x, y)$ is called *algebraic distance* of the point $(x, y)$. To find best ellipse that fits to a set of pixels $(x_i, y_i), i = 1, 2, .., N$, sum of squared algebraic distances is minimized.

$MIN \sum_{i=1}^{N} F(x_i, y_i)^2$        (6)

By applying fitting methodology proposed by Halir et al. in [9], optimal solution of equation (6) can be found by finding eigenvector $a_1 = [a, b, c]^T$ of matrix $M$ with minimal positive eigenvalue $\lambda$, where

$$M a_1 = \lambda a_1$$

$$M = C_1^{-1}(D_1^T D_1 - D_1^T D_2 (D_2^T D_2)^{-1}(D_1^T D_2)^T)$$    (7)

$$a_1^T C_1 a_1 = 1$$

$$a = \begin{pmatrix} a_1 \\ a_2 \end{pmatrix}$$

$$a_2 = [d, e, f]^T = -(D_2^T D_2)^{-1}(D_1^T D_2)^T a_1$$

$$C_1 = \begin{pmatrix} 0 & 0 & 2 \\ 0 & -1 & 0 \\ 2 & 0 & 0 \end{pmatrix}$$

$$D_1 = \begin{pmatrix} x_1^2 & x_1 y_1 & y_1^2 \\ \cdot & \cdot & \cdot \\ x_i^2 & x_i y_i & y_i^2 \\ \cdot & \cdot & \cdot \\ x_N^2 & x_N y_N & y_N^2 \end{pmatrix}$$

$$D_2 = \begin{pmatrix} x_1 & y_1 & 1 \\ \cdot & \cdot & \cdot \\ x_i & y_i & 1 \\ \cdot & \cdot & \cdot \\ x_N & y_N & 1 \end{pmatrix}$$

After finding best ellipse that fits to a set of pixels, code is generated to represent its ellipse arc as $(x_0, y_0, a, b, \emptyset, \beta, \gamma)$, where $(x_0, y_0)$ is the center of the ellipse, a, b are major and minor axises, $\emptyset$ is anticlockwise angle of rotation from x-axis to the major axis (range of $\emptyset$ is from 0 to 180), $\beta$ is the start angle of the arc, and $\gamma$ is the end angle of the arc. As shown in Fig. 9, set of pixels are represented as anticlockwise arc from start to end points.

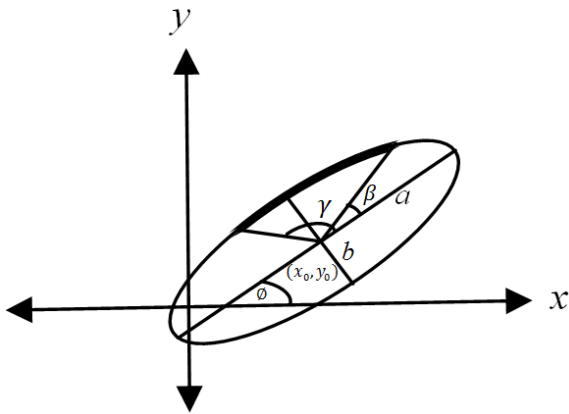

Fig. 9.  Ellipse arc with code $(x_0, y_0, a, b, \emptyset, \beta, \gamma)$

To calculate $\beta$ and $\gamma$, center point of ellipse is moved to origin. $\beta'$ and $\gamma'$ are calculated, where $\beta'$ is anticlockwise angle of rotation from x-axis to the line from origin to start point, and $\gamma'$ is anticlockwise angle of rotation from x-axis to the line from origin to end point. Finally, $\beta$ and $\gamma$ are calculated as $\beta = \beta' - \emptyset$ and $\gamma = \gamma' - \emptyset$.

Table 3 shows examples of detected ellipse arcs from sub-words in Fig. 7. For simplicity, numbers in Table 3 are rounded to nearest integers.

Code for sequence of connected characters or sub-characters is generated as $(c_1, c_2, .., c_i, ..., c_n)$, where $c_i = (code_i, F_{i,1}, F_{i,2}, F_{i,3})$, $code_i$ is point , line, or ellipse code, $F_{i,j}$ is Freeman direction to $c_{i+j}$. Fig. 10 shows example of subwords code. Null direction is represented using (9). Finally, completed code of word or list of words is generated as shown in Fig. 11.

### B. Matching method

To extract characters' code and to recognize characters in text image, the following matching method is applied. Two line segments $(p_i, \propto_i, l_i)$ and $(p_j, \propto_j, l_j)$ are compared by comparing their vectors in Polar form. $(p_i, \propto_i, l_i) \equiv (p_j, \propto_j, l_j)$ if $(l_i, \propto_i) \equiv (l_j, \propto_j)$ . $(l_i, \propto_i)$ is considered equivalent to $(l_j, \propto_j)$ if $|l_i - l_j| < \Delta l$ , $|\propto_i - \propto_j| < \Delta \propto$ , where $\Delta l$ , and $\Delta \propto$ are accuracy factors of length and directions of vectors. All parallel line segments with the same length are equivalent. $(p_i, \propto_i, l_i) \subseteq (p_j, \propto_j, l_j)$ if $(l_i, \propto_i) \subseteq (l_j, \propto_j)$. $(l_i, \propto_i)$ is considered subset from $(l_j, \propto_j)$ if $l_i \leq l_j$, $|\propto_i - \propto_j| < \propto_a$.

TABLE III.    Examples Of Detected Ellipses Arcs And Their Codes

| Ellipse Arcs | code |
|---|---|
| 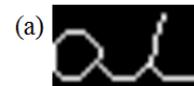 | (19, 10, 275, 0, 0,153,187) <br> (13, 15, 9, 7, 50, 315, 36) <br> (17, 4, 5, 4, 120, 128, 60) |
| | (16, 43, 102, 0, 39,125,302) |
| 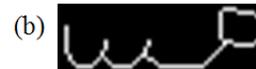 | (16, 62, 6, 4, 119, 339, 31) <br> (6, 72, 16, 9, 165, 334, 24) <br> (-333, -109, 412, 41, 75, 33, 35) |

(a)

(((19, 10, 275, 0, 0,153,187), 7, 4, 0),
((13, 15, 9, 7, 50, 315, 36), 3, 1, 0),
((17, 4, 5, 4, 120, 128, 60), 0, 0, 9),
((27, 54, 12), 7, 9, 9),
((21, 171, 9), 9, 9, 9))

(b)

(((16, 62, 6, 4, 119, 339, 31), 0, 0, 0),
((6, 72, 16, 9, 165, 334, 24), 1, 0, 4),
((77, 14, 2), 6, 4, 7),
((23, 81, 3), 4, 0, 0),
((51, 118, 11), 0, 0, 0),
((29, 12, 14), 1, 1, 1),
((82, 141, 1), 2, 1, 1),
((98, 89, 10), 1, 0, 9),
((17, 137, 11), 6, 9, 9),
((-333, -109, 412, 41, 75, 33, 35), 9, 9, 9))

Fig. 10.  Examples of sub-word and its code

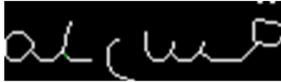

(((((19, 10, 275, 0, 0,153,187), 7, 4, 0),((13, 15, 9, 7, 50, 315, 36), 3, 1, 0),((17, 4, 5, 4, 120, 128, 60), 0, 0, 9),((27, 54, 12), 7, 9, 9),((21,171,9),9,9,9)),0,0,0 ), ((((35, 118, 11), 2, 7, 9),((16, 43, 102, 0, 39, 125, 302), 6, 9, 9), ((13,124,6),9,9,9)),0,1,1 ),((((16, 62, 6, 4, 119, 339, 31), 0, 0, 0),((6, 72, 16, 9, 165, 334, 24), 1, 0, 4),((77, 14, 2), 6, 4, 7),((23,81, 3), 4,0,0 ),((151,118,11),0,0,0 ),((29,12,14),1,1,1 ),((82,141, 1), 2,1,1 ),((98,89,10),1,0,9),((17,137, 11), 6,9,9),((-333, -109 ,412, 41,75,33,35),9,9,9 )),1, 1, 9 ),(((( 0,0,1),9,9,9)),0,9,9 ),((((0,0,1),9,9,9)),9,9,9 ))

Fig. 11. Example of completed word's code

Two ellipse arcs $(x_i, y_i, a_i, b_i, \emptyset_i, \beta_i, \gamma_i)$ and $(x_j, y_j, a_j, b_j, \emptyset_j, \beta_j, \gamma_j)$ are considered equivalent if $|a_i - a_j| < \Delta a$, $|b_i - b_j| < \Delta b$, $\emptyset_i = \emptyset_j$, $\beta_i = \beta_j$, and $\gamma_i = \gamma_j$, where $\Delta a$, and $\Delta b$ are accuracy factors. Ellipse arc $(x_i, y_i, a_i, b_i, \emptyset_i, \beta_i, \gamma_i)$ is considered sub-arc from $(x_j, y_j, a_j, b_j, \emptyset_j, \beta_j, \gamma_j)$ if $|a_i - a_j| < \Delta a$, $|b_i - b_j| < \Delta b$, $\emptyset_i = \emptyset_j$, $\beta_i \geq \beta_j$, $\gamma_i \leq \gamma_j$.

Two codes $(c_1, c_2, .., c_i, ..., c_n)$ and $(d_1, d_2, ..., d_j, ..., d_m)$ are considered equivalent if $n = m$, $c_i \equiv d_i$ $\forall$ i = 1, .., n.

$c_i = (code_i, F_{i,1}, F_{i,2}, F_{i,3})$ is equivalent to $c_j = (code_j, F_{j,1}, F_{j,2}, F_{j,3})$ $if\, code_i \equiv code_j$, $F_{i,1} = F_{j,1}$, $F_{i,2} = F_{j,2}$, and $F_{i,3} = F_{j,3}$.

$c_i = (code_{ci}, F_{ci,1}, F_{ci,2}, F_{ci,3})$ is called matched with $d_{q+k} = (code_{dq+k}, F_{dq+k,1}, F_{dq+k,2}, F_{dq+k,3})$ if $c_i \subseteq (code_{dq+k}, \sum_{r=1}^{k} F_{dq+r,1}, \sum_{r=1}^{k} F_{dq+r,2}, \sum_{r=1}^{k} F_{dq+r,3})$, and $c_{i-1} \subseteq d_q$ or $c_{i-1}$ matches $d_q$, where $\sum_{r=1}^{k} F_{dq+r,1}$ is vector that represents sum of vectors $F_{dq+1,1}, F_{dq+2,1}, ..., F_{dq+k,1}$.

$(c_1, c_2, .., c_i, ..., c_n) \subseteq (d_1, d_2, .., d_j, ..., d_m)$ if $n \leq m$, $\exists d_j \ni c_1 \subseteq d_j$, $and \forall c_i \exists d_r \ni c_i \subseteq d_r$ or $c_i$ matches $d_r$.

### C. Codebook generation

For each font, codebook contains a unique code for each character form as well as a unique list of codes that represents fingerprint of the font.

To identify unique code for each character form, character forms have to be collected first. However, each character may have different forms in the same location in the same font. Therefore, for each location, list of all possible combination of connected characters with maximum length three are generated. This number of characters has been chosen because some characters at the beginning of word change their forms depend on neighbors of their neighbors. Sub-words in each list are classified and code is generated for each class.

Table 4 shows connectivity of Arabic characters that are used to generate sub-words. New character KASHEEDA (-) is added, which can be used in many places in Arabic words. As shown in Table 4, there are 36 character with right connectivity and 25 with left connectivity. Table 5 shows number of sub-words that can be generated for each location.

Code is identified for each sub-word by generating a set of images using different font sizes. Each image is converted to code and common code is extracted to represent sub-word code. The reason behind using different sizes is that, although, character form is not affected by font size in most of existing fonts, thinning process is affected by font size.

Fig. 12 shows the word AKALA (أكل) using different sizes of *Times New Roman* font. As shown in Fig. 12, although, LAMON (ﻟ) character has the same form in the original word with different font sizes, LAMON (ﻟ) character has different thinned forms. In Fig. 12, Zhang-Suen thinning algorithm is applied during thinning process. Same result was reached even with different thinning algorithms (such as Guo-Hall thinning algorithm). Thinning Arabic script has many problems as mentioned in [18]. These problems increase difficulties of generating unique code for each character form using one font size.

TABLE IV. CONNECTIVITY OF ARABIC CHARACTERS

| | character | Right | left | | character | Right | left |
|---|---|---|---|---|---|---|---|
| 1 | - | T | T | 19 | ط | T | T |
| 2 | آ | T | | 20 | ظ | T | T |
| 3 | أ | T | | 21 | ع | T | T |
| 4 | إ | T | | 22 | غ | T | T |
| 5 | ب | T | T | 23 | ف | T | T |
| 6 | ت | T | T | 24 | ق | T | T |
| 7 | ث | T | T | 25 | ك | T | T |
| 8 | ج | T | T | 26 | ل | T | T |
| 9 | ح | T | T | 27 | م | T | T |
| 10 | خ | T | T | 28 | ن | T | T |
| 11 | د | T | | 29 | ه | T | T |
| 12 | ذ | T | | 30 | و | T | |
| 13 | ر | T | | 31 | ؤ | T | |
| 14 | ز | T | | 32 | ى | T | T |
| 15 | س | T | T | 33 | ي | T | T |
| 16 | ش | T | T | 34 | ئ | T | T |
| 17 | ص | T | T | 35 | ء | | |
| 18 | ض | T | T | 36 | ة | T | |

TABLE V. MAXIMUM NUMBER OF GENERATED SUB-WORDS

| Sub-word size | Beginning | Middle | End |
|---|---|---|---|
| 2 | 36 | - | 25 |
| 3 | 25*36 | 25*36 | 25*36 |
| Total | 936 | 900 | 925 |

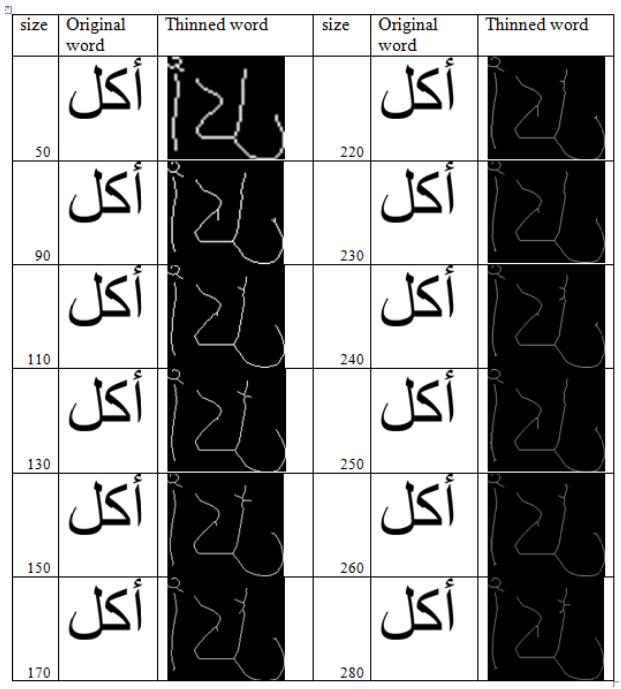

Fig. 12. Thinning problems with different font sizes

| Linear | Ellipse | code |
|---|---|---|
| | | (((((1,76,5),7,0,9 ),((1,150,15),0,9,9 ),((21, 37 ,5, 3,13,144,330),9,9,9 )),6,9,9 ),((34,30),9, 9,9 )) |
| | | (((((1,78,3),5,7,0 ),((1,0,10),0,0,9 ),((1,164, 22),0,9,9),((25, 53 ,6, 5,34,115,344),9,9,9 )),2, 1,9 ),((18,44),0,9,9 ),((18,49),9,9,9 )) |
| | | (((((1,12,14),7,1,1 ),((38, 37 ,5, 1,48,55,247 ),3,2,1 ),((1,78,5),1,1,1 ),((1,82,13),7,0,7 ),((1,99,15),1,6,7 ),((1,0,3),6,6,9 ),((1,121,11), 1,9,9 ),((30, 50 ,71, 0,0,12,196),9,9,9 )),7,9,9 ),((33,41),9,9,9 )) |
| | | (((((1,14,13),1,7,9 ),((1,166,6),6,9,9 ),((-511, 209 ,575, 70,126,342,343),9,9,9 )),2,9,9 ),((17,34),9,9,9 )) |
| | | (((((31, 29 ,16, 8,117,201,243),7,0,7 ),((1, 146,14),1,0,0 ),((1,46,21),7,0,6 ),((1,0,1), 3,3,4 ),((1,14,2),4,7,0 ),((1,88,2),7,0,9 ),((1,126,9),1,9,9 ),((1,62,11),9,9,9 )),9,9,9 )) |
| | | (((((1,78,1),6,6,6 ),((1,39,1),0,1,1 ),((1,40, 11),2,2,1 ),((1,103,13),1,7,9 ),((22, 14 ,17, 7,91,69,117),6,9,9 ),((1,83,6),9,9,9 )),9,9,9 )) |
| | | (((17,27),4,6,9 ),((17,22),6,9,9 ),((((29, 21 ,6, 4,87,180,343),6,5,7 ),((1,92,2),2,7,0 ),((1,78,14),7,0,9 ),((1,150,15),0,9,9 ),((37, 38 ,8, 7,99,4,95),9,9,9 )),9,9,9 )) |
| | | (((((31, 39 ,7, 3,154,222,321),5,5,6 ),((1,78,1 ),3,7,1 ),((1,0,9),7,0,1 ),((1,133,10),2,2,1 ),((50,106,16),1,1,7 ),((1,153,9),7,6,9 ),((1,169, 3),6,9,9 ),((41, 37 ,8, 6,101,25,152),9,9,9 )),2,9,9 ),((1,0,1),9,9,9 )),9,9,9 )) |

Fig. 13. Examples of generated characters' code in *Tahoma* font

| Linear | Ellipse | code |
|---|---|---|
| | | (((((1,75,13),7,0,9 ),((1,8,29),1,9,9 ),((768, 235 ,796, 4,46,194,194),9,9,9 )),6,9,9 ),((((-96, -10 ,127, 2,38,11,12),9,9,9 )),9,9,9 )) |
| | | (((((1,72,12),7,0,0 ),((1,0,34),0,1,9 ),((13, 36 ,6, 4,24,141,349),2,9,9 ),((1,166,2),9,9,9 )),2, 9,9 ),((((2, 23 ,4, 1,179,49,237),0,9,9 ),((1,78, 1),9,9,9 )),9,9,9 )) |
| | | (((((-94, -5 ,119, 2,38,11,11),9,9,9 )),2,9,9 ),((26, 12 ,130, 0,78,257,64),2,2,2 ),((4, 15 ,194, 0,0,14,206),6,5,1 ),((1,141,1),5,1,6 ),((1,11, 15),1,7,0 ),((1,0,18),6,7,7 ),((1,48,14),2,1,9 ),((1,60,14),7,9,9 ),((17, 30 ,5, 1,128,174,344), 9,9,9 )),9,9,9 )) |
| | | (((((1,166,2),4,2,7 ),((1,0,5),1,7,7 ),((1,30, 13),7,7,9 ),((1,60,13),0,9,9 ),((23, 60 ,58, 0, 125,285,285),9,9,9 )),2,9,9 ),((((-113, -10 ,123, 2,38,11,11),9,9,9 )),9,9,9 )) |
| | | (((((8, 6 ,7, 5,108,181,270),7,0,0 ),((1,167, 17),1,1,0 ),((1,71,19),0,7,7 ),((1,86,10),6,7,7 ),((1,165,8),7,0,0 ),((1,147,14),0,1,9 ),((1, 141,5),2,9,9 ),((1,123,12),9,9,9 )),9,9,9 )) |
| | | (((((33, 3 ,97, 0,78,258,63),1,0,1 ),((1,100,13), 6,1,2 ),((1,40,15),2,2,1 ),((63, -878 ,902, 55, 89,94,95),3,6,6 ),((1,38,6),7,6,9 ),((1,144,10), 6,9,9 ),((1,47,3),9,9,9 )),9,9,9 )) |
| | | (((((2, 6 ,1, 4,77,50,239),0,9,9 ),((1,78,1),9,9 ,9 )),6,9,9 ),((((1,77,15),6,1,7 ),((29, 1 ,201, 0, 141,50,281),2,7,0 ),((15, 5 ,4, 3,0,214,325),6, 7,7 ),((1,153,13),0,1,9 ),((31, 21 ,3, 1,39,129, 322),2,9,9 ),((1,125,12),9,9,9 )),9,9,9 )) |

Fig. 14. Examples of generated characters' code in *Times New Roman* font

## IV. EVALUATION

The proposed technique has been implemented using *OpenCV*. To implement linear and ellipse regressions, Geometric Regression Library (*GeoRegression*) is used. *GeoRegression* is an open source Java geometry library for scientific computing [19].

Codebook is generated for two fonts (*Times New Roman*, and *Tahoma*). Fig. 13 and Fig. 14 show examples of generated characters' code in *Times New Roman* and *Tahoma* fonts. Table 6 shows examples of sub-words that are generated to identify code of character (ط)(in *Tahoma* font and in middle of word). In Table 6, linear and ellipse parts of each sub-word as well as its code are shown. Common parts are detected using the proposed matching technique and exploited to generate character code. Common parts are underlined in Table 6. Fig. 15 shows generated code of character (ط). To illustrate validity of the generated character code, Arabic word (أمطار) is printed in image using font *Tahoma* with size 50. As shown in Fig. 16, code of the word (أمطار) contains code sequence that matches with code of character (ط).

Performance of the proposed recognition technique has been evaluated using 14822 different words, which are collected from Holy Quran. Sequences of words are selected, converted to images, and used as inputs to the proposed technique. The proposed technique has recognized 12750 words with *Tahoma* font and 12350 words with *Times New Roman* font. Which means that recognition rate of the proposed technique is 86% in *Tahoma* font and 83% in *Times New Roman* font.

TABLE VI.    EXAMPLES OF GENERATED SUB-WORDS TO IDENTIFY CHARACTER (ط) CODE IN TAHOMA FONT

| size | Linear parts | Ellipse parts | code |
|------|--------------|---------------|------|
| 50 | | | (((((26,144,4),1,0,3 ),((36,103,5),5,4,4 ),((31,95,9),3,4,7 ),((22,47,16),5,7,0 ),((19,92,18),0,0,0 ),((31,177,46),2,1,0 ),((6,85,15),1,7,7 ),((57,102,31),7,7,7 ),((32,0,1),7,1,1 ),((18,130 ,6),2,2,2 ),((75,84,12),1,1,7 ),((0,134,18),1,6,7 ),((32,39,14),6,7,7 ),((65,53,12),1,1,9 ),((36, 94 ,927, 0,0,289,164),2,9,9 ),((75,149,9),9,9,9 ),9,9,9 )) |
| 75 | | | (((((19, 5 ,316, 0,78,97,282),6,5,0 ),((32, 4 ,8, 0,124,261,107),4,1,0 ),((21,39,1),1,0,2 ),(( 31,103,5),5,4,5 ),((22,95,9),4,4,7 ),((4,73,9),0,0,0 ),((5,133,12),0,0,0 ),((31,176,41),2,1,0 ),((1,85,15),1,7,0 ),((48,102,31),7,7,7 ),((32,0,2),1,0,0 ),((65,78,9),7,7,0 ),((21,153,13),0,1,0 ),((-91, -4 ,148, 9,106,32,34),2,0,7 ),((78,78,2),7,7,0 ),((35,6,12),7,1,7 ),((82,124,5),2,7,7 ),((97,70,7),7,7,7 ),((10,127,12),0,1,1 ),((73,55,3),1,1,9 ),((120,26,9),2,9,9 ),((107,115,4),9 ,9,9 )),9,9,9 )) |
| 100 | | | ((((41,0,3,0),1,1,9 ),((((24,103,5),5,5,7 ),((9,95,9),4,7,1 ),((2,98,11),0,0,0 ),((32,177,42),2,1 ,0),((4,85,15),1,7,7 ),((35,102,31),7,7,7 ),((-697, -474 ,49, 898,10,36,36),7,7,7 ),((46,104,5) ,7,0,1 ),((23,129,13),0,1,1 ),((64,67,8),1,1,9 ),((76,78,1),6,9,9 ),((33, 75 ,4, 0,153,180,348) ,9,9,9 )),0,9,9 ),((19.0,65,0),9,9,9 )) |
| 120 | | | (((((24,103,5),5,4,7 ),((10,95,9),3,7,1 ),((0,81,31),7,0,0 ),((32,177,43),2,1,0 ),((4,85,15),1,0 ,7 ),((36,102,31),7,7,7 ),((53,78,5),7,0,0 ),((27,166,27),0,0,9 ),((53,53,9),1,9,9 ),((-600, 204 ,642, 3,147,349,349),9,9,9 )),7,9,9 ),((41.0,68,0),9,9,9 )) |
| 50 | | | (((((34,103,5),5,4,5 ),((29,2,6),2,0,3 ),((3,8,8),7,3,5 ),((28,95,9),3,4,7 ),((16,36,12),6,7,0 ),((28,151,15),0,0,0 ),((32,178,53),1,1,0 ),((4,85,15),1,0,7 ),((54,102,31),7,7,7 ),((71,78,5) ,7,0,0 ),((26,166,27),0,0,9 ),((58,53,9),1,9,9 ),((-600, 222 ,642, 3,147,349,349),9,9,9 )),7 ,9,9 ),((41.0,86,0),9,9,9 )) |
| 75 | | | (((((28, 20 ,1, 0,113,33,213),0,3,1 ),((28,78,1),3,1,0 ),((18,88,2),0,0,7 ),((38,103,5),5,5,4 ),((35,95,9),5,4,4 ),((34,29,7),4,3,4 ),((4,74,9),0,7,0 ),((10,74,17),6,0,0 ),((30,156,21),0,0,0 ),((31,177,39),2,1,0 ),((8,85,15),1,0,0 ),((61,102,31),7,7,7 ),((17,104,4),6,5,6 ),((81,143,11) ,4,7,7 ),((73,143,18),7,0,0 ),((20,152,20),0,0,9 ),((70,62,8),1,9,9 ),((30, 114 ,41, 0,0,167, 343),9,9,9 ),7,0,9 ),((44.0.98.0),0,9,9 ),((44.0,105.0),9,9,9 )) |
| 100 | | | (((38.0,8.0),0,9,9 ),((((26,144,4),1,0,3 ),((36,103,5),5,4,4 ),((31,95,9),3,4,7 ),((22,47,16) ,5,7,0 ),((19,92,18),0,0,0 ),((31,177,43),1,1,0 ),((6,85,15),1,7,7 ),((57,102,31),7,7,0 ),((29, 79 ,5, 3,101,49,266),0,2,4 ),((82,141,2),3,4,3 ),((30,26,5),5,5,7 ),((57,79,4),1,0,9 ),((77,23, 11),0,9,9 ),((69,151,21),9,9,9 )),9,9,9 )) |
| 120 | | | (((((34,103,5),5,4,5 ),((29,2,6),2,0,3 ),((3,8,8),7,3,5 ),((28,95,9),3,4,7 ),((16,36,12),6,7,0 ),((28,151,15),0,0,0 ),((32,178,54),1,1,0 ),((4,85,15),1,7,7 ),((54,102,31),7,7,7 ),((-59, 538 ,478 , 21,140,280,280),3,2,2 ),((32,0,1),1,1,1 ),((72,84,12),1,1,7 ),((0,134,18),1,6,7 ),((32,39,14), 6,7,7 ),((64,53,12),1,1,9 ),((98,36,3),2,9,9 ),((73,149,9),9,9,9 )),9,9,9 )) |
| 50 | | | (((6.0,10.0),7,0,9 ),((((34,103,5),5,4,5 ),((29,2,6),2,0,3 ),((3,8,8),7,3,5 ),((28,95,9),3,4,7 ),(( 16,36,12),6,7,0 ),((28,151,15),0,0,0 ),((32,178,53),1,1,0 ),((4,85,15),1,7,7 ),((54,102,31), 7, 7,7 ),((-539, -341 ,43, 704,10,36,37),7,7,7 ),((65,104,5),7,0,1 ),((18,129,13),0,1,1 ),((71, 67, 8),1,1,9 ),((95,78,1),6,9,9 ),((33, 94 ,4, 0,153,348,180),9,9,9 )),0,9,9 ),((19.0,84.0),9,9,9 )) |
| 75 | | | (((2.0,8.0),7,0,0 ),((((13, 3 ,3, 0,39,134,317),7,6,0 ),((21,141,1),5,0,7 ),((25,92,2),1,0,1 ),(( 28,103,5),5,3,4 ),((17,95,9),3,4,4 ),((1,34,8),5,5,7 ),((23,47,10),4,7,0 ),((3,107,14),0,0,0 ),(( 32,178,47),1,1,0 ),((0,85,15),1,0,7 ),((43,102,31),7,7,7 ),((60,78,5),7,0,0 ),((26,166,27),0,0, 9),((55,53,9),1,9,9 ),((-600, 211 ,642, 3,147,349,349),9,9,9 )),0,0,9 ),((13.0.73.0),0,9,9 ),(( 13.80),9,9,9 )) |
| 100 | | | (((2.0,11.0),4,7,9 ),((2.0,4.0),7,9,9 ),((((13, 3 ,3, 0,39,134,317),7,6,0 ),((21,141,1),5,0,7 ),((25,92,2),1,0,1 ),((28,103,5),5,3,4 ),((17,95,9),3,4,4 ),((1,34,8),5,5,7 ),((23,47,10),4,7,0 ),((3,107,14),0,0,0 ),((32,178,47),0,1,1 ),((32,0,3),3,3,1 ),((0,85,15),1,0,7 ),((43,102,31) ,0,7,7 ),((60,100,38),7,7,9 ),((23,135,14),0,9,9 )),((35, 78 ,6, 4,134,140,342),9,9,9 )),9,9,9 )) |
| 120 | | | (((9.0,0.0),0,1,7 ),((9.0,8.0),3,7,7 ),((4.0,4.0),7,7,9 ),((((24,103,5),5,5,7 ),((10,95,9),4,7,1 ),((1,98,11),0,0,0 ),((32,177,42),2,1,0 ),((4,85,15),1,0,7 ),((36,102,31),7,7,7 ),((53,78,5), 7,0,0 ),((27,166,27),0,0,9 ),((53,53,9),1,9,9 ),((41.0,68.0),9,9,9 )) |
| 50 | | | (((((28, 20 ,1, 0,113,33,213),0,3,1 ),((28,78,1),3,1,0 ),((18,88,2),0,0,7 ),((38,103,5),5,5,4 ),((35,95,9),5,4,4 ),((34,29,7),4,3,4 ),((4,74,9),0,7,0 ),((10,74,17),6,0,0 ),((30,156,21),0,0,0 ),((31,177,41),2,1,0 ),((8,85,15),1,7,0 ),((61,102,31),7,7,7 ),((32,0,2),1,0,0 ),((78,78,9),7,7,0 ),((19,153,13),0,1,0 ),((-75, 19 ,128, 8,106,32,34),2,0,0 ),((91,78,2),7,0,7 ),((35,6,12),1,7,7 ),((28, 106 ,76, 0,0,350,172),6,7,7 ),((94,124,5),7,7,0 ),((6,127,12),0,1,1 ),((41, 121 ,73, 0, 102,18,97),1,1,9 ),((132,26,9),2,9,9 ),((120,115,4),9,9,9 )),9,9,9 )) |
| 75 | | | (((2.0,11.0),4,7,7 ),((2.0,4.0),7,7,9 ),((((13, 3 ,3, 0,39,134,317),7,6,0 ),((21,141,1),5,0,7 ),(( 25,92,2),1,0,1 ),((28,103,5),5,3,4 ),((17,95,9),3,4,4 ),((1,34,8),5,5,7 ),((23,47,10),4,7,0 ),((3, 107,14),0,0,0 ),((32,178,48),1,1,0 ),((0,85,15),1,7,7 ),((43,102,31),7,7,7 ),((-59, 526 ,476, 21,140,280,280),3,2,2 ),((32,0,1),1,1,1 ),((61,84,12),1,1,7 ),((2,134,18),1,6,7 ),((29,39,14), 6,7,7 ),((61,53,12),1,1,9 ),((36, 80 ,53, 0,0,295,164),2,9,9 ),((63,149,9),9,9,9 )),0,9,9 ),((33, 70),9,9,9 )) |

| font | Location | Linear | Ellipse | code |
|------|----------|--------|---------|------|
| Tahoma | middle | | | ((1,103,5),5,7,0 ), ((1,95,9),7,1,1 ), ((1,177,43),2,1,9 ), ((1,85,15),1,9,9 ), ((1,102,31),9,9,9 ) |

Fig. 15. Generated code of (ط) character

| Linear | |
|--------|--|
| Ellipse | |
| code | (((((2,78,33),9,9,9 )),2,0,0 ),((((2, 2 ,3 ,3,48,230,336),9,9,9 )),7,7,9 ),((((31, 18 ,224, 0,78,97,282),6,5,0 ),((44, 17 ,8, 0,124,261,107),4,1,0 ),((20,39,1),1,0,2 ),((48,103,5),5,4,5 ),((34,95,9),4,4,7 ),((19,73,9),0,0,0 ),((13,133,12),0,0,0 ),((43,176,41),2,1,0 ),((1,85,15),1,7,0 ),((61,102,31),7,0,9 ),((44,0,2),2,9,9 ),((78,95,31),9,9,9 )),0,9,9 ),((((78,111,13), 2,7,9 ),((96,14,2),7,9,9 ),((52, 94 ,4, 0,29,63,241),9,9,9 )),9, 9,9 )) |

Fig. 16. Code of (أمطار) word contains code of (ط) character

## V. CONCLUSION

Although, extensive researches have been done in recognizing printed text in different languages, only a few have been done in recognizing printed Arabic characters due to complexity of its characters. This paper has proposed recognition technique to recognize printed Arabic text using linear and ellipse regression techniques. Characters are recognized by using their codes, which are sequences of points, lines, and ellipses. Characters' forms are collected by generating all possible combination of three characters for each character location. Common code is extracted from generated sub-words for each character location and stored as code. To differentiate between fonts, fingerprint of each font is identified by collected codes that uniquely exist in this font.

As future work, the proposed recognition technique will be examined with most of existing fonts to evaluate its performance with different fonts as well as with multi-font texts. Moreover, optimization technique will be exploited to optimize accuracy factors used in the proposed technique.